\def\datasetName{TPC--268}

\documentclass[10pt,twocolumn,letterpaper]{article}

\usepackage[T1]{fontenc}

\usepackage[pagenumbers]{cvpr} 

\definecolor{cvprblue}{rgb}{0.21,0.49,0.74}
\usepackage[pagebackref,breaklinks,colorlinks,allcolors=cvprblue]{hyperref}
\usepackage{tabularx}
\usepackage{multirow}
\usepackage{booktabs} 
\usepackage{longtable} 
\usepackage{arydshln}   
\usepackage{ltablex}
\usepackage{xltabular}
\usepackage{xcolor}
\usepackage{supertabular}
\usepackage{xspace}

\definecolor{myred}{HTML}{D62728}
\definecolor{myblue}{HTML}{1F77B4}

\usepackage[skip=5pt]{caption}
\def\paperID{} 
\def\confName{CVPR}
\def\confYear{2026}

\title{Plant Taxonomy Meets Plant Counting: A Fine-Grained, Taxonomic Dataset for Counting Hundreds of Plant Species}

\author{
Jinyu Xu, Tianqi Hu, Xiaonan Hu, Letian Zhou, Songliang Cao, Meng Zhang, Hao Lu\thanks{Corresponding author.} \\
Huazhong University of Science and Technology, China \\
{\tt\small \{jinyu\_xu, hlu\}@hust.edu.cn}
}

\begin{document}
\maketitle
\begin{abstract}
Visually cataloging and quantifying the natural world 
requires pushing the boundaries of both 
detailed visual classification and counting at scale. Despite significant progress, particularly in crowd and traffic analysis, the fine-grained, taxonomy-aware plant counting remains underexplored in vision. 
In contrast to crowds, plants 
exhibit nonrigid morphologies and 
physical appearance variations across 
growth stages and environments. To
fill this gap, we present \datasetName{}, the  
first plant counting benchmark incorporating plant taxonomy. 
Our dataset couples instance-level point annotations with 
Linnaean labels (kingdom$\rightarrow$species) and organ categories, enabling hierarchical reasoning and species-aware 
evaluation. The dataset
features $10,000$ images with $678,050$ point annotations, includes $268$ countable plant categories over $242$ plant species in Plantae and Fungi, and spans observation scales from canopy-level 
remote sensing imagery to tissue-level microscopy.
We follow the problem setting of class-agnostic counting (CAC), provide taxonomy-consistent, scale-aware data splits, and benchmark state-of-the-art regression- and detection-based CAC approaches. By capturing the biodiversity, hierarchical structure, and multi-scale nature of botanical and mycological taxa, \datasetName{} provides a biologically grounded testbed to advance 
fine-grained class-agnostic counting. Dataset and code are available at \url{https://github.com/tiny-smart/TPC-268}.
\end{abstract}
\section{Introduction}
\label{sec:intro}

High-quality datasets such as PASCAL VOC~\cite{everingham2010pascal}, ImageNet~\citep{deng2009imagenet}, and COCO~\citep{lin2014microsoft} have driven vision studies in classification, detection, and segmentation for decades.
A 
similar line 
also appears in visual counting~\cite{chan2008privacy}, a task  
aiming to count objects in 
extreme clutter and occlusion. 
Interestingly this field is dominated by the study of rigid objects like crowds~\cite{zhang2016single} and vehicles~\cite{hsieh2017drone}. Consequently, most 
technical improvements are overfitted into these types of objects, 
rendering poor generalization when counting other 
categories in the natural world, plants for instance. 

\begin{figure}
    \centering
    \includegraphics[width=\linewidth]{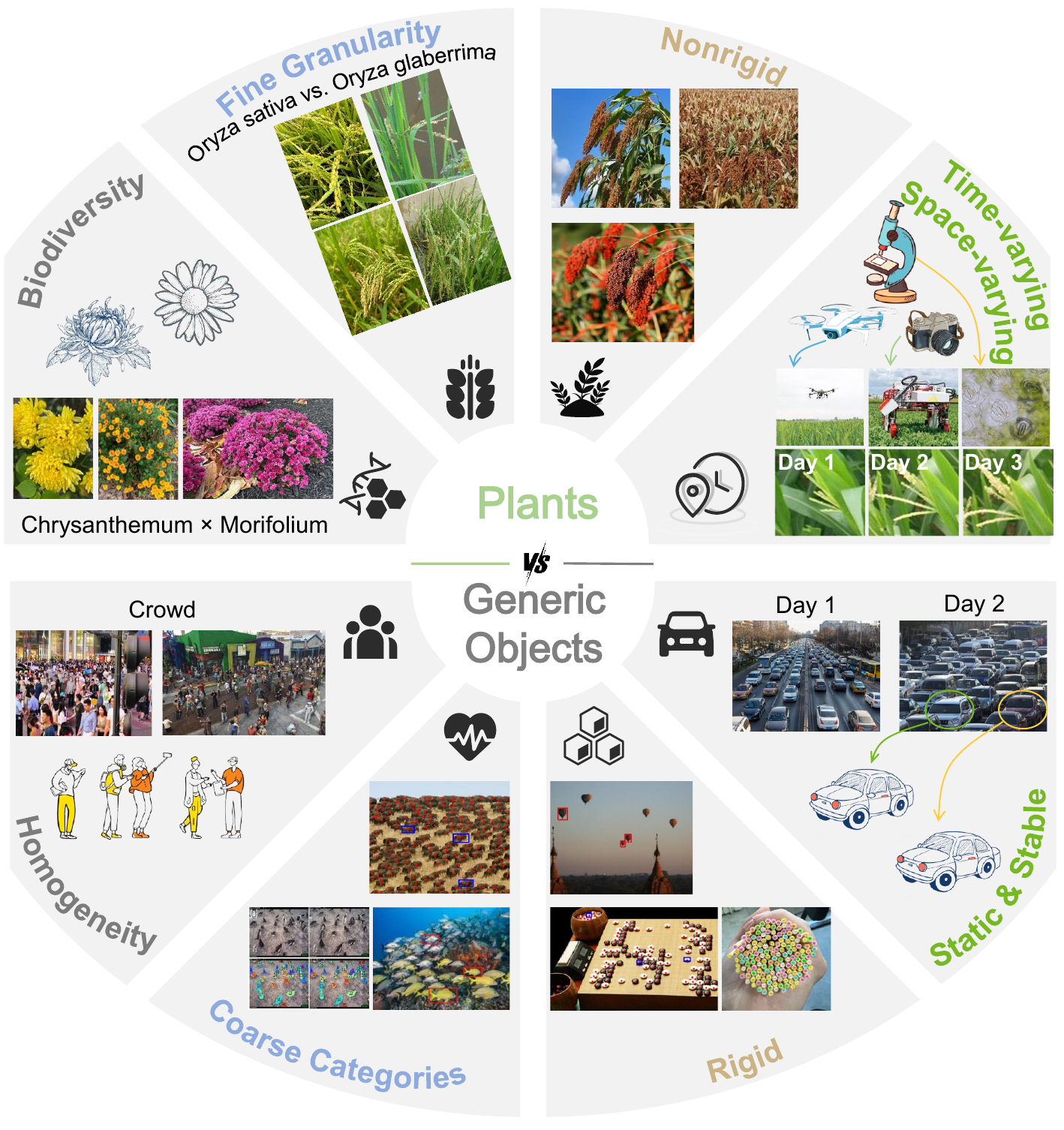}
    \caption{\textbf{Counting plants versus counting generic objects}. 
    Plants inherently exhibit rich biodiversity, fine-grained variations, non-rigid structures, and time-space variations, which collectively create a nonnegligible gap 
    against generic objects.}
    \label{fig:visual_charateristics}
\end{figure}

Plant counting is not merely a niche application of existing counting 
approaches~\cite{lu2017tasselnet}; it represents a fundamentally distinct and more demanding set of challenges. As shown in Fig.~\ref{fig:visual_charateristics}, unlike crowds, where individuals often share 
similar appearance that can provide informative contextual cues~\citep{Cho_2023_CVPR}, plants exhibit staggering diversity and complexity. They are non-rigid, undergo dramatic morphological changes during their life cycle, and 
is highly sensitive to environmental factors---a concept known as phenotypic plasticity~\cite{whitman2009phenotypic}. This 
task is thus not simple single-class counting, but \textit{fine-grained instance counting across a vast and hierarchically structured class space}. For example, 
a crowd counting model 
only requires distinguishing ``person'' from ``background'', but a plant counting system must learn the subtle textural differences to separate hundreds of species, a problem more akin to fine-grained visual categorization~\cite{inat2018} but at a massive scale of instance density. This unique intersection of large-scale counting and fine-grained, taxonomy-aware recognition has been largely overlooked by the vision community. A possible reason may be 
due to the lack of a suitable benchmark. 

\begin{figure}
    \centering
    \includegraphics[width=\linewidth]{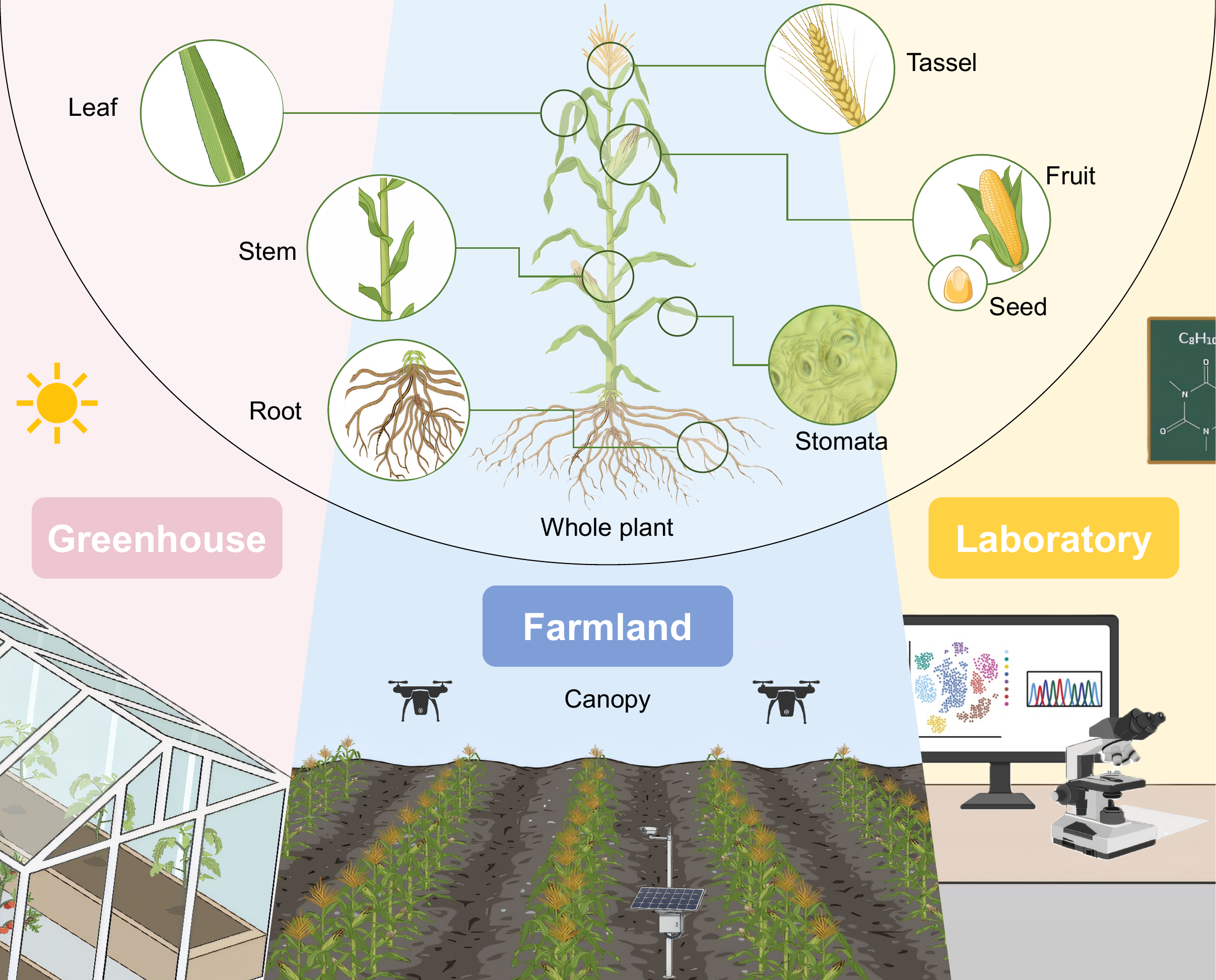}
    \caption{\textbf{Distinct organization-level countable plant categories across multiple observation scales and environments in the \datasetName{} dataset.} Our \datasetName{} covers four organization levels including tissue, organ, organism, and population, spans across various observation scales (ranging from microscopy and close-range photography to UAV imagery), and hosts $268$ countable categories (such as leaves, stems, and fruits) under heterogeneous environments (from laboratory to field).}
    \label{fig:multiscale}
\end{figure}

To this end, we introduce \datasetName{}, the first large-scale, plant-orientated counting dataset that explicitly integrates 
plant taxonomy. Substantially larger than prior plant counting datasets introduced in plant science~\cite{david2020gwhd,lu2017tasselnet,li2023leafcount,fetter2019stomatacounter,hani2020minneapple}, \datasetName{} comprises $10,000$ images, featuring $678,050$ point and $30,000$ bounding box annotations. It covers a remarkable diversity of life, including $242$ plant species, organized into $268$ distinct biological organization-level countable categories (\eg, different organs of the same species are treated as separate counting categories). A key feature of this dataset is its deep integration of plant taxonomy. Unlike generic objects, plants possess inherent, systematic priors encoded in their evolutions. We annotate each instance with its full taxonomic hierarchy (from kingdom to species), transforming the counting problem into \textit{joint counting and hierarchical reasoning}. This rich, structured information allows the vision community to investigate not only how visual representations learned at one taxonomic level (\eg, family) can generalize to unseen species within it, but also how shared visual traits correspond to phylogenetic closeness. This explicit encoding of biological structure provides a principled framework for developing robust and generalizable category-agnostic counting models. Furthermore, we provide multi-level annotations (Fig.~\ref{fig:multiscale}) that decompose plants into differing organizations (\eg, tissues, organs, whole plants, and canopies), enabling the study of cross-species understanding of homologous structures and model generalization across vast changes in observation scale, from macroscopic canopies to microscopic stomata.

The sheer taxonomic scale of the plant kingdom renders the conventional 'one-model-per-species' paradigm fundamentally unscalable. This reality motivates us to frame plant counting as a Class-Agnostic Counting (CAC) problem, where the goal is to learn the general concept of ``how to count'' given visual exemplar(s). 
Doing so not only aligns with the practical need of plant science but also establishes a new, complex frontier for CAC research, which necessitates a principled methodological and experimental design to navigate. A core principle of our dataset design is a strict, taxonomy-aware evaluation of generalization. Unlike existing benchmarks, our data splits are defined at the species-organization level, guaranteeing that a model tested on, for example, an ``Oryza sativa leaf'' has never seen any leaf from the entire Oryza genus during training. This methodology allows us to rigorously benchmark how a model generalizes across a true, biologically-defined taxonomic gap---a far more challenging and realistic measure of zero-shot counting. Under this strict protocol, we provide a comprehensive benchmark of state-of-the-art CAC approaches, systematically analyzing how different architectural priors tackle the unique challenges posed by our dataset, from fine-grained distinctions to extreme scale variations (Fig.~\ref{fig:multiscale}). By grounding the class-agnostic challenge in a biologically meaningful hierarchy and providing this rigorous evaluation framework, \datasetName{} serves as a new, challenging testbed to spur innovation in fine-grained visual counting.

\begin{table*}[t!] 
\small
\centering
\caption{\textbf{Representative visual counting datasets introduced by the computer vision community in the past ten years}.}
\label{tab:datasets}
\renewcommand{\arraystretch}{1}
\begin{tabular*}{\textwidth}{@{}ll@{\extracolsep{\fill}}ccccc@{}}
\toprule
Dataset & Venue & \#Images & \#Labeled Instances & Avg. Resolution & Target & \#Classes\\
\midrule
ShanghaiTech A~\cite{zhang2016single} & CVPR'16 & 482 & 241,677 & 868$\times$589 & Crowd &  1\\
ShanghaiTech B~\cite{zhang2016single} & CVPR'16 & 716 & 88,488 & 1024$\times$768 & Crowd & 1 \\
Penguin dataset~\cite{arteta2016counting} & ECCV'16 & 80,095 & 575,082 &1920$\times$1080 & Penguin & 1\\
CARPK~\cite{hsieh2017drone} & ICCV'17 & 1,448 &  89,777 &1280$\times$720& Car  & 1 \\
CVPPP~\cite{Dobrescu_2017_ICCV} & ICCVW'17 & 1,311 & 18,016 & 842$\times$812& Leaf  & 1 \\
DCC~\cite{marsden2018people} & CVPR'18  & 177 & 6,036 &256$\times$256 & Cell & 1 \\
UCF--QNRF~\cite{idrees2018composition} & ECCV'18  & 1,535 &1,251,642& 2902$\times$2013 & Crowd & 1 \\
JHU--CROWD++~\cite{sindagi2020jhu} & TPAMI'20 & 4,375 & 1,515,005 &1430$\times$910& Crowd & 1 \\
NWPU--Crowd~\cite{gao2020nwpu} &  TPAMI'20 & 5,109 &2,133,375  & 3209$\times$2191 & Crowd & 1 \\
IOCfish5K~\cite{sun2023ioc} & CVPR'23 & 5,637 & 659,024 & 1920$\times$1080& Fish & 1 \\
\midrule
FSC--147~\cite{famnet} & CVPR'21 & 6,135 & 344,150  &523$\times$384& Miscellaneous& 147 \\
FSCD--LVIS~\cite{nguyen2022few} & ECCV'22 & 6,196 & 402,945 &586$\times$479& Miscellaneous& 377 \\
Mara--Wildlife~\cite{kumar2024wildlifemapper} & CVPR'24 & 1,012 & 28,146 & 8256$\times$5504 & Animal & 21 \\
\datasetName{} & This work & 10,000 & 678,050 &1130$\times$959& Plant & 268  \\
\bottomrule
\end{tabular*}
\end{table*}

\section{Related Work}
\label{sec:formatting}

Our work is related to single-image counting datasets, plant counting, and class-agnostic counting.

\vspace{-10pt}
\paragraph{Single-Image Counting Datasets.}

Counting datasets have evolved from class-specific to class-agnostic 
ones.
Table.~\ref{tab:datasets} compares main counting datasets introduced in the vision community over the past decade. It can be observed that
early work~\cite{zhang2016single,idrees2018composition,sindagi2020jhu} focuses on a single category in dense scenarios, primarily targeting crowds where severe occlusions and perspective changes pose major challenges. Subsequent work expands to other domains such as cells~\cite{marsden2018people}, animals~\cite{arteta2016counting,sun2023ioc}, leaves~\cite{minervini2016finely}, and vehicles~\cite{hsieh2017drone}.
A limitation of these datasets is the expensive annotation cost, as a single image can contain thousands of annotations, impeding scalability to new scenarios and categories.
This bottleneck motivates a paradigm shift toward CAC~\cite{lu2018class}, which reframes the objective from learning what to count to learning how to count. However, the practical success of this paradigm hinges on the diversity and granularity of the underlying benchmarking data. Existing CAC datasets~\cite{famnet,nguyen2022few} lack the fine-grained categorical complexity found in many real-world domains. While effort such as WildLife-Mapper~\cite{kumar2024wildlifemapper} has begun to address this gap for wildlife counting, 
the more challenging plant kingdom remains underexplored. To date, the community still lacks a large-scale, fine-grained counting dataset.

\vspace{-10pt}
\paragraph{Plant Counting.} 
Plant counting is a statistical approach primarily used and developed by experts in plant science and agriculture to link genotypes with plant phenotypes~\cite{furbank2011phenomics}. It 
underpins various agricultural tasks, including emergence rate, biomass and yield estimation in breeding trials. Early methods, mostly driven by the plant science community, relied on sensors~\citep{Luck2008SensorRT, ehsani2009two}, digital image processing techniques~\citep{gnadinger2017digital} with tools like ImageJ~\citep{schneider2012nih} and SmartGrain~\citep{tanabata2012smartgrain}. Yet, these approaches suffer from reliance on manual feature engineering and are sensitive to environmental variations. 
Only until recently, deep-learning based approaches are introduced in plant counting~\citep{giuffrida2016learning, bargoti2017deep, lu2017tasselnet}, significantly accelerating the iteration of plant counting techniques.
Albeit effective, these approaches can only count specific plant species. Once the counting category changes, a repetitive loop of data collection, annotation, and model retraining is required. 
This tedious workflow yields a collection of multiple yet isolated species-specific plant counting datasets. 
To address this limitation and catalyze more generalizable plant counting approaches, we construct a novel plant counting dataset, TPC--268, featuring a hierarchical taxonomic structure for plant-agnostic counting.

\vspace{-10pt}
\paragraph{Class-Agnostic Counting.}
Conventional counting methods can only count predefined categories, such as crowds~\cite{zhang2016single}, vehicles~\cite{hsieh2017drone}, and cells~\cite{he2021deeply}. To mitigate this category dependency, 
CAC is introduced to generalize a counting model for unseen categories. CAC is a typical exemplar-to-image semantic correspondence problem~\cite{shi2022represent,you2023few,lin2022scale,liu2022countr,wang2024vision}. With several predefined exemplars as input, they are matched with the image to search visually similar content, and the density map is used to encode the object count. 
To improve output interpretability, \cite{nguyen2022few} marries CAC with detection. 
This detection-based paradigm directly locates and counts instances with bounding boxes. Besides these exemplar-guided approaches, 
a text-guided paradigm also emerges, which uses text prompts for open-set object counting~\cite{liu2025countse}. Across these paradigms, while the conditioning signal may differ, their goal remains the same: counting instances of a user-specified, novel concept at test time. In this work, we extend the problem connotation of CAC in the plant counting domain and curate a plant counting benchmark with point-level annotations to 
assess generalization across taxonomic and fine-grained plant species.


%
\section{Taxonomic Plant Counting Dataset}

Here we introduce our benchmark, \datasetName{}, a large-scale taxonomic plant counting dataset, present 
its hierarchical taxonomic structure, 
and 
provide its statistics.

\subsection{Plant Taxonomy Meets Plant Counting}
We highlight plant taxonomy for plant counting. 
Plant taxonomy organizes species as a nest hierarchical ranks: \texttt{Kingdom}, \texttt{Phylum}, \texttt{Class}, \texttt{Order}, \texttt{Family}, \texttt{Genus}, and \texttt{Species}. In this structure, higher-level ranks (\eg, class, order, family) group plants that share broad morphological and ecological patterns such as growth form, leaf and stem architecture, or reproductive structures, while lower-level ranks (\eg, genus and species) distinguish subtler phenotypic differences among closely related taxa, providing a structured similarity potential for visual cues. Building on this observation, we make counting explicitly taxon-aware by embedding each counting instances into the Linnaean hierarchy~\cite{linnaeus1789systema}, rather than treating the species as unrelated categoric labels. 
The annotation can thus be interpreted and aggregated at multiple taxonomic levels (\eg, species-level vs. genus- or family-level counts), and category splits can be defined in terms of taxonomic distance.
With this taxonomic annotation available, a model can be encouraged to generalize from seen specie to unseen but taxonomic related species.

\begin{figure}
    \begin{minipage}{\columnwidth}
        \centering
        \includegraphics[width=\columnwidth]{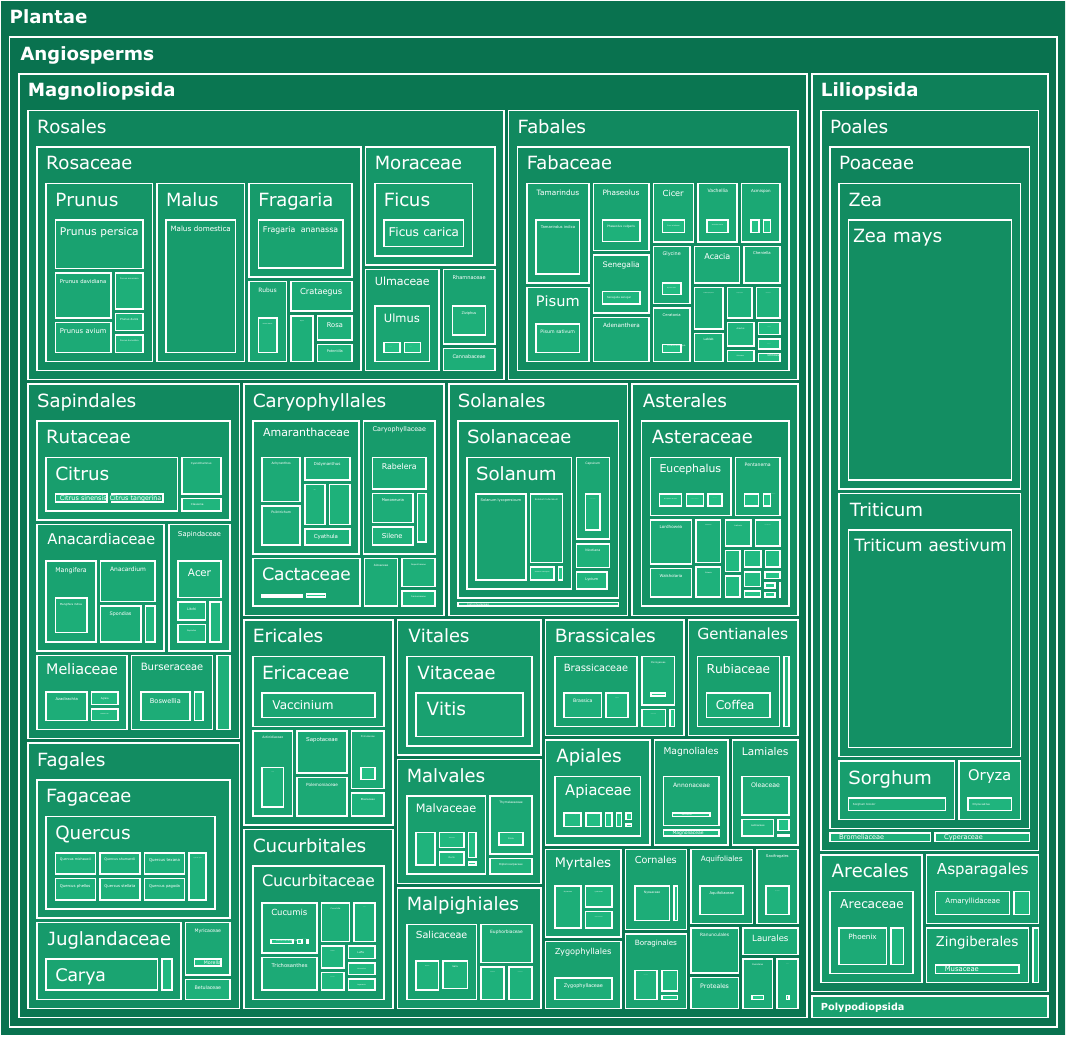}
        \caption{\textbf{Treemap of plant taxonomic hierarchy in TPC-268.} Each nested rectangle represents a specific taxonomic rank, from Kingdom (Plantae), Phylum (\eg, Angiosperms), Class (\eg, Magnoliopsida, Liliopsida), Order (\eg, Rosales, Poales), Family (\eg, Rosaceae, Poaceae), Genus (\eg, Prunus, Zea), to Species (\eg, Prunus persica, Zea mays). Box area represents the number of images for that taxonomic unit.}
        \label{fig:placeholder}
    \end{minipage}
\end{figure}

\subsection{Dataset Collection}

\begin{figure*}
    \centering
    \includegraphics[width=\linewidth]{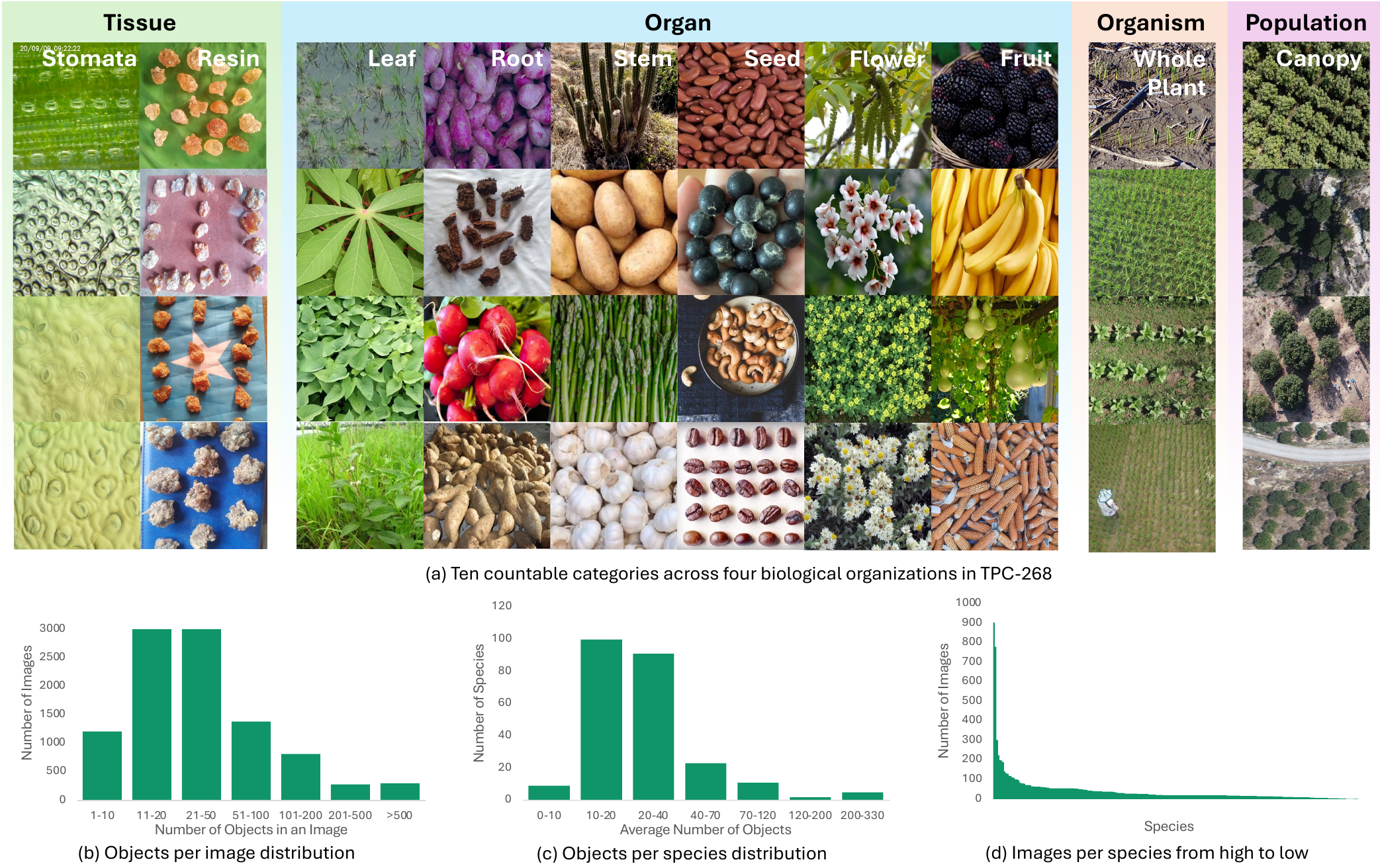}
    \caption{\textbf{Examples and statistical distributions of \datasetName{}.} The figure illustrates (a) representative samples of ten countable plant categories across four biological organizational levels. The associated statistical distributions show (b) the number of objects per image, (c) the number of objects per species, and (d) the number of images-per-species. 
}
    \label{fig:dataset_statics}
\end{figure*}

The dataset 
integrates curated public resources with controlled data collection to achieve broad taxonomic coverage and data reliability.
Major sources are distributed among Wikipedia (34\%), \textit{PlantCLEF} (29\%)~\cite{martellucci2025overview}, Internet (14\%), Tree Leaf Stomata (11\%)~\cite{Wang2024Stomata}, \textit{MTC-UAV} (3\%)~\cite{lu2021tasselnetv3}, \textit{CVPPP} (3\%)~\cite{Dobrescu_2017_ICCV}, \textit{CKC-Wild} (3\%)~\cite{cornpheno}, and others (3\%).
All data undergoes a rigorous preprocessing pipeline to filter out samples unsuitable for counting, with annotations being manually refined or newly created to ensure quality.

The dataset spans scales from tissue-level microscopy to canopy-level UAV imagery. Collected across laboratories, greenhouses, and natural habitats, the images include variations in illumination, background, and occlusion, alongside common degradations such as blur and compression.

The dataset encompasses extensive geographical and ecological diversity across Asia, Africa, \etc covering ecosystems from tropical to alpine environments. It incorporates major crops such as \textit{Oryza sativa}, and \textit{Zea mays}, alongside specialized arid-zone plants like \textit{Acacia tortilis}.

The dataset also incorporates ecologically rare, geographically restricted, and conservation-significant taxa. Examples include the critically endangered \textit{Arenaria paludicola}, near-threatened \textit{Boswellia sacra}, and regionally endemic \textit{Jacobaea taitungensis} and \textit{Nubelaria arisanensis}. These species are validated against the IUCN Red List~\cite{list2011iucn}. 

The deliberate inclusion introduces a naturally long-tailed distribution across both taxonomic and morphological dimensions, providing a valuable testbed for studying model robustness and hierarchical generalization under extreme sample imbalance and ecological rarity.

Image sources comply with copyright licenses. Sensitive geographical metadata is anonymized and used solely for ecological stratification to protect privacy.

\subsection{Annotation Protocols}

Our annotation strategy is designed to capture both the instance location for counting and the taxonomic identity for hierarchical analysis.

\vspace{-10pt}
\paragraph{Annotation of Instance and Exemplar.} 
 
We adopt the standard point-box protocol in CAC~\cite{famnet}, providing point annotations at the structural center of each instance and bounding boxes for three instances per image. For partial occlusion, we follow a \textit{visible-shape dominance} principle, annotating only visible structures. Box exemplars are selected to cover: i) representative appearance, ii) diverse scales, and iii) 
morphological or environmental variations.

\vspace{-10pt}
\paragraph{Annotation of Taxonomy and Biological Organization.}
A key feature of our dataset is its complete taxonomic hierarchy. Each image is linked to the full Linnaean taxonomy~\cite{linnaeus1789systema} of its primary species.
In practice, we used \textit{Pl@ntNet}~\cite{joly2016look} for initial genus-species identification, completed the remaining hierarchy using the \textit{World Flora Online} database~\cite{world_flora_online}, and subjected all data to rigorous manual verification. Annotators cross-referenced identified species with \textit{Pl@ntNet}’s organ-specific images; all labels underwent a 3-round human check.
Finally, each species' position is encoded as a $7$-dimensional vector. A complete mapping table can be found in the supplementary. For example, \textit{Malus domestica} is encoded as $[1, 1, 1, 14, 39, 113, 136]$.

\begin{figure}
    \centering
    \includegraphics[width=\linewidth]{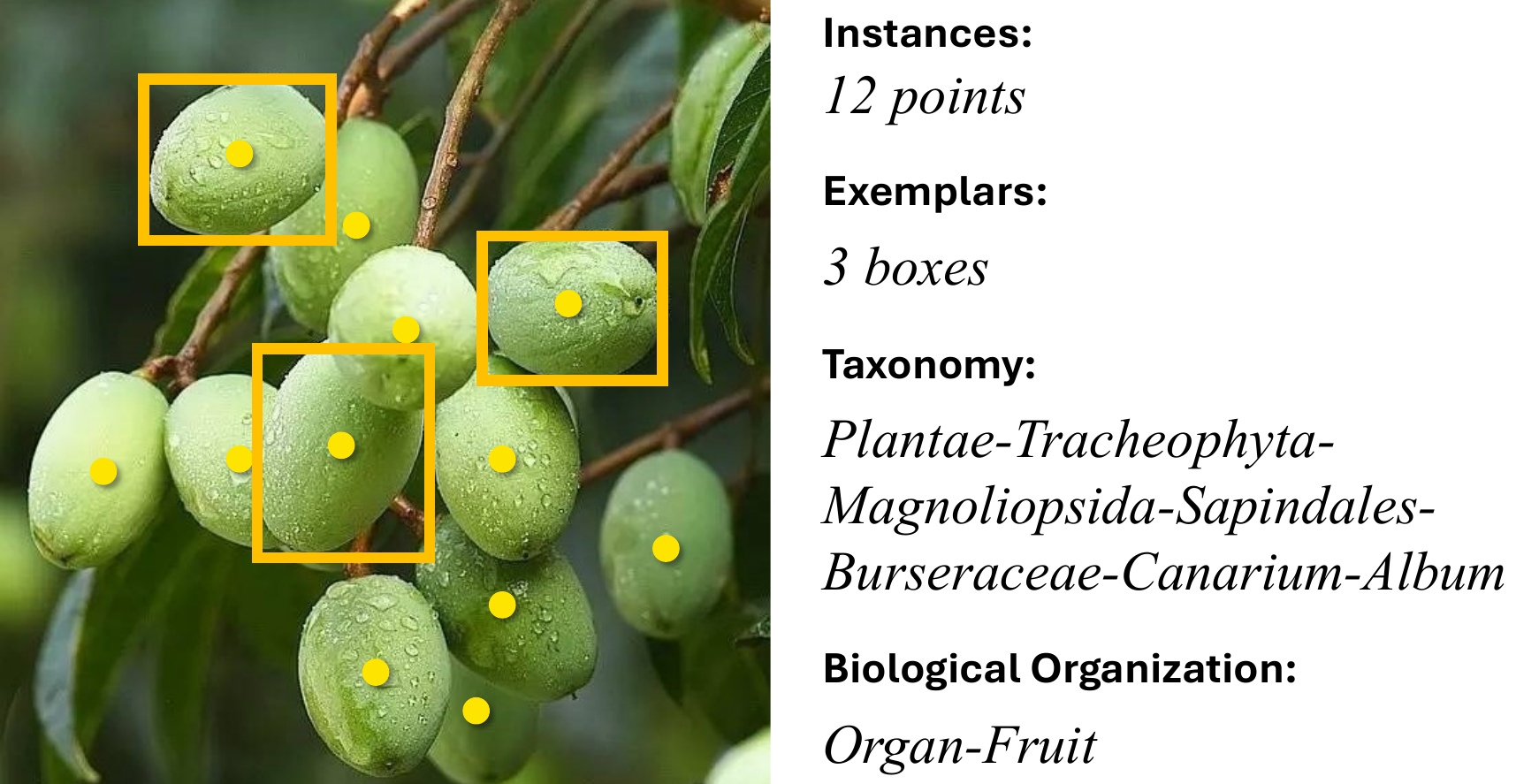}
    \caption{\textbf{An annotation example}. Each sample contains annotations for instances, exemplars, taxonomy, and organization.}
    \label{fig:dataset_annotation}
\end{figure}

Biological organization information (\eg, stomata, flower, and whole plant) is also included as auxiliary metadata, and its distribution across ten major categories is detailed in Fig.~\ref{fig:dataset_statics}(a). This metadata specifies the type of biological structure depicted in the image.
Specifically, for multi-organ handling, images with distinct multiple organs (e.g., flowers and fruits) are annotated as separate categories with specific suffixes, whereas indiscernible cluttered regions are treated as background.
This label is critical for vision tasks, as the visual features (\eg, texture, shape, color) of the same species vary dramatically depending on the structure shown (\eg, a Malus domestica ``leaf'' image versus its ``whole plant'' image).
This information thus helps to horizontally categorize different plant varieties based on the specific biological structure depicted, not just by species.
Fig.~\ref{fig:dataset_annotation} shows an example of annotation.

\vspace{-10pt}
\paragraph{Annotation Pipeline.} 
Our annotation pipeline combines public data sources with newly created labels. 
Annotations for approximately $80$\% of the images were created from scratch by a professional annotation team over three months. For the remaining images with existing bounding boxes, we extracted box centers as points. Subsequently, all annotations (new and sourced) undergo a rigorous three-round review by researchers experienced in plant phenotyping to ensure accuracy, particularly for complex cases like overlapping organs and dense regions.

\subsection{Dataset Partition}

To enforce strict category independence between subsets, we partition the dataset based on taxonomic identity and visual characteristics (\eg, observation scale and density).

The minimal indivisible unit (category) is defined as a species–organization pair (\eg, \textit{Triticum aestivum}--flower vs. \textit{Triticum aestivum}--stomata). Each pair is treated as an independent category and is assigned 
to one subset ({\small \textsf{train}, \textsf{val}, or \textsf{test}}), preventing any instance overlap between different sets.
We formulate the partition as a multi-objective optimization problem and solve it using a Mixed-Integer Linear Programming (MILP) model with three constraints:
i) scale coverage: each split must include at least one instance 
from every observation scale (microscopy, close-up, and remote sensing); ii) density balancing: the average number of points per image should maintain balanced across all subsets;
and iii) ratio adherence: an approximate $7$:$1$:$2$ split for training, validation, and testing.

This split achieves a balanced average density of $67.81$ instances per image across all subsets. 
The MILP-based approach ensures that, while the splits are taxonomically distinct, they remain statistically balanced in 
instance density and observation scale coverage. 
Further details w.r.t. the MILP-based partition can refer to the supplementary. 

\subsection{Statistics and Analysis}
The TPC-268 dataset contains $10,000$ images, including annotations of $678,050$ points and $30,000$ bounding boxes.

\vspace{-10pt}
\paragraph{Distribution of Density.}
Instance counts range from $5$ to $1,462$ with a long-tailed distribution (median: $25.0$; $75$-th percentile: $58.0$; $90$-th percentile: $129.0$). $72.1\%$ of images contain fewer than $50$ instances, $22.0\%$ contain between $50$ and $200$, and $3.0\%$ exceed $500$, as visualized in Fig.~\ref{fig:dataset_statics}(b).

\vspace{-10pt}
\paragraph{Hierarchical Taxonomic Structure.}
The dataset follows the Linnaean system, including $2$ kingdoms, $2$ phyla, $4$ classes, $35$ orders, $83$ families, $192$ genera, and $242$ species. Long-tailed distributions persist across all ranks. At the family level, the mean image count ($121.9$) exceeds the median ($40$). Ranked by image count, the top $10$ families, genera, and species account for $60.9\%$, $38.9\%$, and $32.0\%$ of images, respectively. $53.5\%$ of species contain fewer than $20$ images, and $8.3\%$ contain $5$ or fewer. This species-level skew is illustrated in Fig.~\ref{fig:dataset_statics}(c) and Fig.~\ref{fig:dataset_statics}(d).

\vspace{-10pt}
\paragraph{Biological Organization and Scale.}
The dataset spans four organizational levels: tissue ($1,096$ stomata, $228$ resin), organ ($4,422$ fruit, $2,994$ flower, $602$ seed, $196$ stem, $118$ root, $74$ leaf), organism ($214$ whole plant), and population ($56$ canopy). Observation scales range from microscopic structures ($165 \times 127$ pixels) to macroscopic canopy-level drone imagery ($6,000 \times 4,000$ pixels) . Fig.~\ref{fig:dataset_statics}(a) provides the corresponding example images.

\begin{table*}[t]
    \centering
    \caption{\textbf{Comparison with the state-of-the-art CAC approaches on the \datasetName{} dataset.} Best performance is in \textbf{boldface}.}
    \label{tab:baseline.metrics}
    \small
    \renewcommand{\arraystretch}{1.1}
    \addtolength{\tabcolsep}{-1pt}
    \begin{tabular*}{\textwidth}{@{}llcccccccc@{}}
        \toprule
        
        \multirow{2}{*}{Method} & \multirow{2}{*}{Venue \& Year} & \multirow{2}{*}{Backbone} & \multirow{2}{*}{Shot} 
        & \multicolumn{3}{c}{Val} & \multicolumn{3}{c}{Test} \\ 
        \cmidrule(l){5-7} \cmidrule(l){8-10}
          &    &   &   & MAE$\downarrow$ & RMSE$\downarrow$ & $R^2\uparrow$ & MAE$\downarrow$  & RMSE$\downarrow$   &  $R^2\uparrow$\\
        \midrule 
        FamNet~\cite{famnet}       & CVPR'21   & R50   & 3 & 28.87  & 52.51  & 0.58  & 30.43  & 65.62  & 0.62\\ 
        BMNet+~\cite{shi2022represent}       & CVPR'22  & R50    & 3 & 29.33  & 77.78  & 0.47  & 27.78  & 57.25   & 0.50\\ 
        C-DETR~\cite{nguyen2022few} & ECCV'22 & R50 & 3 & 22.66 & 77.51 & 0.75 & 22.68 & 57.97 & 0.74\\ 
        SPDCNet~\cite{lin2022scale}      & BMVC'22  & R18    & 3 & 25.66  & 72.49  & 0.52  & 23.70  & 47.53  & 0.64\\
        CountTR~\cite{liu2022countr}      & BMVC'22  & Hybrid     & 3 & 20.21  & 55.82  & 0.73  & 25.19  & 49.94  & 0.62\\ 
        SAFECount~\cite{you2023few}    & WACV'23  & R18     & 3 & 22.57  & 63.65  & 0.64  & 25.70  & 52.30  & 0.58\\ 
        LOCA~\cite{djukic2023low}         & ICCV'23 &  R50 & 3 & 17.26  & 53.19  & 0.75   & \textbf{17.51}   & \textbf{38.37}   & \textbf{0.78}\\ 
        DAVE~\cite{pelhan2024dave}          & CVPR'24 &  R50  & 3 &16.47  & 52.87  & 0.76   & 17.61   & 40.06   & 0.75\\
        CACViT~\cite{wang2024vision}       & AAAI'24  & ViT-B    & 3 & 16.63  & \textbf{42.49}  & 0.82 & 22.04  & 41.79  & 0.73\\
        CountGD~\cite{amini2024countgd} & NeurIPS'24 & Swin-B & 3 & 18.32 & 54.55 & 0.74 & 19.52 & 50.51 & 0.61\\ 
        TasselNetV4~\cite{hu2025tasselnetv4}  & ISPRS'26 & ViT-B & 3 & \textbf{13.20}  & 43.93  & \textbf{0.83}  & 22.95  & 51.36  & 0.60\\ 
        \midrule
        FamNet~\cite{famnet}        & CVPR'21 & R50   & 1 & 33.11{\footnotesize $\pm$0.68} & 68.95{\footnotesize $\pm$4.15} & 0.58{\footnotesize $\pm$0.05} & 33.63{\footnotesize $\pm$1.13} & 62.07{\footnotesize $\pm$2.94} & 0.41{\footnotesize $\pm$0.05} \\  
        BMNet+~\cite{shi2022represent}& CVPR'22 & R50   & 1 & 29.33{\footnotesize $\pm$0.13}   & 77.50{\footnotesize $\pm$0.31}   & 0.48{\footnotesize $\pm$0.05}    & 27.84{\footnotesize $\pm$0.10} & 56.98{\footnotesize $\pm$0.12} & 0.50{\footnotesize $\pm$0.01} \\   
        CountTR~\cite{liu2022countr} & BMVC'22 & Hybrid  & 1 & 20.16{\footnotesize $\pm$0.05}   & 55.15{\footnotesize $\pm$0.82}   & 0.73{\footnotesize $\pm$0.01}    & 25.19{\footnotesize $\pm$0.14} & 50.23{\footnotesize $\pm$0.24} & 0.62{\footnotesize $\pm$0.00} \\
        LOCA~\cite{djukic2023low} & ICCV'23 & R50 & 1 & 17.19{\footnotesize $\pm$0.31}   & 48.14{\footnotesize $\pm$2.19}   & 0.80{\footnotesize $\pm$0.02}    & 21.47{\footnotesize $\pm$0.29} & \textbf{42.36}{\footnotesize $\pm$0.72} & \textbf{0.73}{\footnotesize $\pm$0.01} \\
        DAVE~\cite{pelhan2024dave} & CVPR'24 &  R50 & 1 &16.06{\footnotesize $\pm$0.60}   & 48.35{\footnotesize $\pm$1.19}   & 0.80{\footnotesize $\pm$0.01}    & \textbf{19.47}{\footnotesize $\pm$0.44} & 42.54{\footnotesize $\pm$0.35} & 0.72{\footnotesize $\pm$0.00} \\
        CACViT~\cite{wang2024vision} & AAAI'24  & ViT-B   & 1 & 17.96{\footnotesize $\pm$0.16}  & 43.38{\footnotesize $\pm$0.47}  & 0.83{\footnotesize $\pm$0.00}    & 22.06{\footnotesize $\pm$0.11} & 42.97{\footnotesize $\pm$0.81} & 0.71{\footnotesize $\pm$0.01} \\ 
        TasselNetV4~\cite{hu2025tasselnetv4}  & ISPRS'26 & ViT-B &1 & \textbf{13.49}{\footnotesize $\pm0.02$}   & \textbf{41.30}{\footnotesize $\pm0.46$}    & \textbf{0.85}{\footnotesize $\pm0.00$}   & 22.20{\footnotesize $\pm0.11$}  & 48.70{\footnotesize $\pm0.26$}  & 0.67{\footnotesize $\pm0.00$} \\

        \bottomrule
    \end{tabular*}
\end{table*}

\section{Results and Discussion}
In this section, we conduct comprehensive experiments to assess the proposed \datasetName{} benchmark from multiple perspectives. 
We benchmark a wide range of representative CAC approaches under a unified evaluation setting, 
encompassing both regression-based and detection-based paradigms.
Furthermore, we investigate cross-dataset transfer to analyze the domain discrepancies between plant counting and generic object counting.

\subsection{Experimental Setup}
We evaluate our dataset under the standard exemplar-based CAC paradigm, where each test image is accompanied by $K$ exemplar patches cropped from the original image. Following existing CAC approaches~\cite{lu2018class}, we consider both the \textit{3-shot} and \textit{1-shot} settings. For cross-dataset transfer, we train models separately on \datasetName{} and FSC--147, and evaluate the performance on another dataset.

We benchmark both regression-based and detection-based CAC frameworks. Regression-based methods follow the exemplar-to-density 
paradigm, where the model takes an image and exemplar prompts as input to predict a density map. We include: 1) FamNet~\cite{famnet}, SAFECount~\cite{you2023few}, and BMNet+\cite{shi2022represent}: CNN-based works that establish the feature matching and density regression framework; 
2) SPDCNet~\cite{lin2022scale}: a CNN-based model utilizing scale-prior deformable convolutions;
3) CountTR~\cite{liu2022countr} and CACViT~\cite{wang2024vision}: transformer-based approaches that leverage self-attention to capture global context; 
4) LOCA~\cite{djukic2023low}: a strong baseline that incorporates local similarity matching and scale-adaptive modules;
5) DAVE~\cite{pelhan2024dave}: a two-stage detect-and-verify framework; 
6) TasselNetV4~\cite{hu2025tasselnetv4}: a vision foundation model for cross-scene, cross-scale, and cross-species plant counting.
Detection-based methods reformulate counting as an exemplar-guided detection task, providing instance-level localization. We consider: 
1) C-DETR~\cite{nguyen2022few}: a DETR-based model that directly predicts bounding boxes for all instances similar to the exemplars; 2) CountGD~\cite{amini2024countgd}: a recent method that uses Gaussian heatmaps for instance localization. This selection ensures a balanced comparison across different architectures (CNN vs. Transformer) and output representations (density maps vs. bounding boxes), providing insights into which paradigm is more suitable for plant counting. 

\begin{table*}[t]
\caption{\textbf{Results of cross-dataset transfer}. \textit{A$\rightarrow$B} denotes model trained on dataset A and tested on the test set of dataset B. \textcolor{myred}{Red}/\textcolor{myblue}{Blue} indicate MAE increase/decrease compared to training and testing on the same dataset (\textit{A$\rightarrow$A}).}
\label{tab:cross-dataset}
\setlength{\tabcolsep}{4.5pt} 
\renewcommand{\arraystretch}{1.1} 
\small
\centering
\begin{tabular*}{\textwidth}{@{}l@{\extracolsep{\fill}}cccccccccccc}
\toprule
\multirow{2}{*}{Method} &
  \multicolumn{3}{c}{FSC--147$\rightarrow$FSC--147} &
  \multicolumn{3}{c}{FSC--147$\rightarrow$\datasetName{}} &
  \multicolumn{3}{c}{\datasetName{}$\rightarrow$\datasetName{}} &
  \multicolumn{3}{c}{\datasetName{}$\rightarrow$FSC--147} \\
  \cmidrule(l){2-4} \cmidrule(l){5-7} \cmidrule(l){8-10} \cmidrule(l){11-13}
       & MAE & RMSE & $R^2$ & MAE & RMSE & $R^2$ & MAE & RMSE & $R^2$ & MAE & RMSE & $R^2$ \\ \midrule
CountTR~\cite{liu2022countr}  & 11.90 & 90.88 & 0.62  & 38.62 \textcolor{myred}{\scriptsize (+225\%)} & 75.78 & 0.12  & 25.19 & 49.94 & 0.62 & 26.53 \textcolor{myred}{\scriptsize (+5\%)} & 126.25 & 0.26 \\
CACViT~\cite{wang2024vision}  & 10.83 & 73.12 & 0.75  & 26.73 \textcolor{myred}{\scriptsize (+147\%)} & 103.35 & -0.63  & 22.04 & 41.79 & 0.73 & 17.88 \textcolor{myblue}{\scriptsize (-19\%)} & 82.57 & 0.68 \\
LOCA~\cite{djukic2023low}   & 10.72 & 56.10 & 0.85  & 24.70 \textcolor{myred}{\scriptsize (+130\%)} & 68.68 & 0.59  & 17.51 & 38.37 & 0.78 & 15.16 \textcolor{myblue}{\scriptsize (-13\%)} & 109.15 & 0.45 \\ \bottomrule
\end{tabular*}
\end{table*}

\subsection{Benchmark Results}
Main results on \datasetName{} are shown in Table.~\ref{tab:baseline.metrics}. Note that each image in our dataset has 3 visual exemplars. For the \textit{1-shot} setup, we randomly select one and report the standard deviation. For \textit{3-shot}, since all exemplars are used, no test-time randomness is introduced, therefore no standard deviation. Overall, regression-based models outperform detection-based models.  This indicates that explicit object localization is hindered by the compact spatial arrangement and structural entanglement present in our dataset.

Within the regression-based models, LOCA achieves the best performance on the test set, surpassing both CNNs and Transformer designs,suggesting that integrating local structure cues with global context modeling effectively captures fine morphological details and dense overlapping present in the dataset. 
In contrast, models relying primarily on global self-attention, such as CACViT and TasselnetV4, demonstrate strong validation performance but generalize poorly to unseen scenes in test. 
Since most baselines show negligible val-test differences, this gap originates from the approach per se rather than data sampling. 
During parameter selection, these global models tend to select hyperparameters that overfit the validation data. TasselNetV4 still demonstrates strong overall performance and higher $R^2$ scores on the validation split, showing that global interaction is beneficial for density estimation. However, its performance gap on the test set suggests that the species variation and structural diversity present in the dataset challenge purely global feature reasoning. This highlights that capturing local structural consistency is essential for robust generalization across species and imaging conditions.

For detection-based approaches, C-DETR and CountGD perform markedly 
worse than regression models. 
Their poor performance arises from distinguishing individual instances 
under severe occlusion and high morphological similarity. 
This indicates that our dataset benefits more from holistic estimation than explicit instance localization.

\begin{figure}[t]
    \centering
    \includegraphics[width=\linewidth]{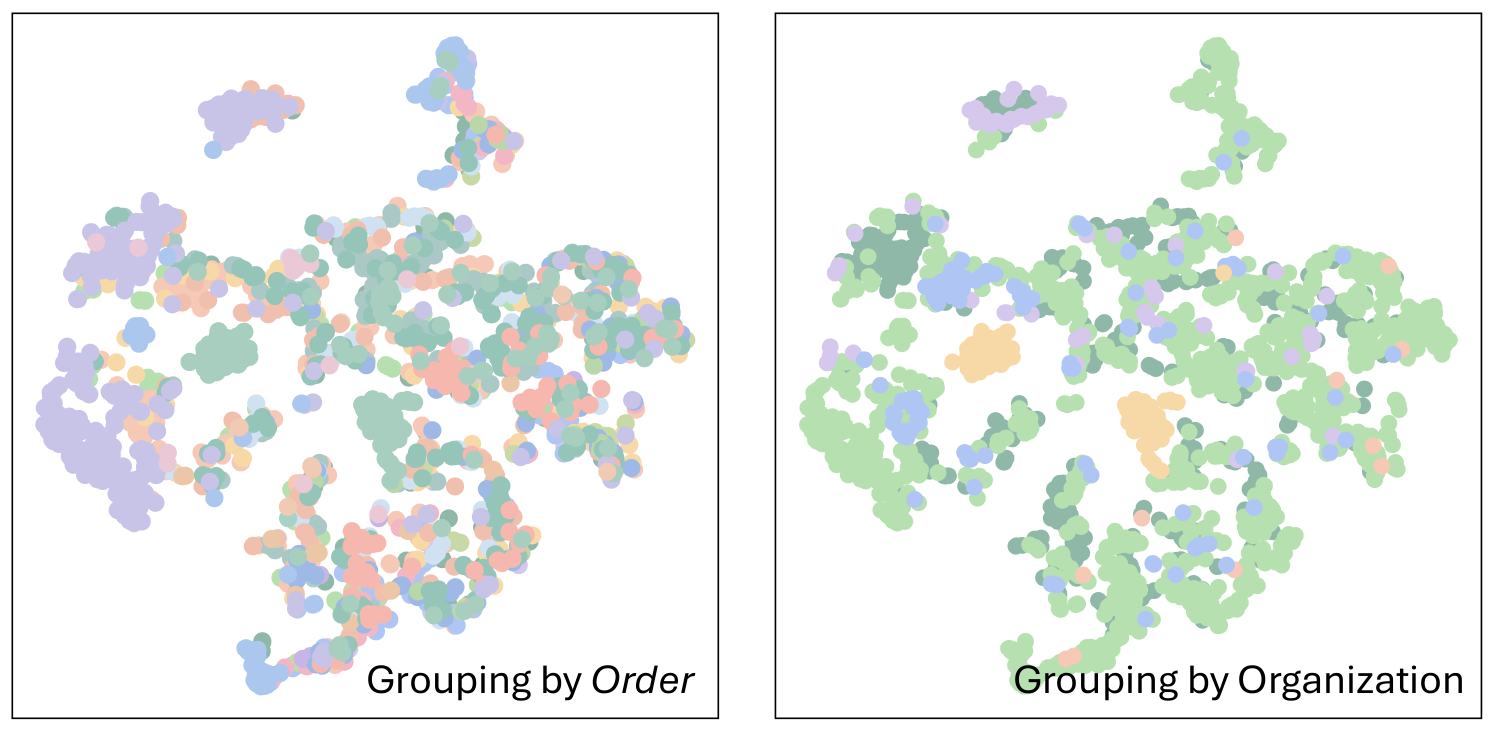}
    \caption{\textbf{t-SNE visualization of LOCA~\cite{djukic2023low} prototype features on the test set.} Different colors indicate different groups.}
    \label{fig:tsne}
\end{figure}

We visualize test samples via t-SNE~\cite{maaten2008visualizing}, using 256-dimensional feature vectors computed as the mean of their LOCA~\cite{djukic2023low} prototypes.
The results are shown in Fig.~\ref{fig:tsne}, where the features are grouped and colored by Order-level taxonomy (left) and organization structure (right). 
Results reveal that 
samples sharing the same label fail to form well-separated clusters under either scheme. 
This 
confirms that the learned features lack clear class-level distinctions, and the SOTA method is insufficient to capture deep biological characteristics when relying solely on visual information.

\begin{table}[t]
\caption{\textbf{Performance of CountGD on the \datasetName{} test set with different prompts.} The species and taxonomic information are provided in text.}
\small
\begin{tabular*}{\linewidth}{@{}l@{\extracolsep{\fill}}ccc}
\toprule
Target Specification  & MAE$\downarrow$ & RMSE$\downarrow$ & $R^2\uparrow$ \\ \midrule
3 visual exemplars                 & 19.52    & 50.51    & 0.61      \\
\hspace{10pt}+ species name  &     17.53&     44.80&       0.69\\
\hspace{10pt}+ full taxonomy    &     16.90&     43.32&       0.71\\ \bottomrule
\end{tabular*}
\label{tab:countgd}
\end{table}

\subsection{Fine-Grained Performance Evaluation}

A fine-grained evaluation of LOCA reveals distinct performance variations across taxonomic, scaling, and data dimensions.
Taxonomic analysis shows that errors in \textit{Brassicaceae} (MAE $62.4$) and \textit{Poaceae} ($54.7$) exceed those in \textit{Rosaceae} ($14.1$), with \textit{Brassica} ($141.5$) and \textit{Zea} ($67.4$) being the primary error sources due to their extreme densities ($228.2$ and $172.3$) and severe occlusion. 
Across observation scales, LOCA excels in microscopic settings (MAE $3.71$) but struggles in macroscopic ($20.30$) and aerial ($15.87$) views. For high-density samples ($>100$), microscopic error remains low ($5.9$) while macroscopic error spikes ($53.7$), implying robustness to repetitive patterns but sensitivity to complex spatial occlusions.
The weak correlation between sample quantity and error (Pearson $r=0.32$) further confirms that counting difficulty is intrinsically tied to morphological complexity rather than data scale.

\subsection{Cross-Dataset Transfer}

Table~\ref{tab:cross-dataset} reports the performance of three models on cross-dataset setting.
A key observation is the substantial performance degradation across all models when trained on FSC--147 and tested on \datasetName{}, indicating that the model trained on generic objects struggle to generalize to plants. 
Conversely, when the training and testing datasets are exchanged, the MAE shows negligible increases or even reductions.
This indicates that our plant counting dataset is a more challenging task than FSC--147 and the model trained on \datasetName{} can generalize to generic objects naturally.
We note that the RMSE and $R^2$ metrics exhibit more pronounced performance degradation, which is expected as these metrics are more sensitive to the large errors and outliers common in challenging domain adaptation scenarios.

\subsection{Insights into Taxonomic Information}

To validate the utility of the proposed hierarchical taxonomic annotations, we retrain CountGD~\cite{amini2024countgd} by incorporating taxonomic information as textual descriptions. 
Compared with the visual-exemplars-only setting, in the species-level experiment we add a text prompt containing the species name. For the full taxonomy setting, the text prompt is in the format: ``\textit{Kingdom, Phylum, Class, Order, Family, Genus, Species.}''
The models are retrained and evaluated under these configurations. 
As shown in Table.~\ref{tab:countgd}, this consistent improvement confirms that structured biological knowledge provides a 
practical inductive bias for the task.

We further evaluate the impact of taxonomy on LOCA using unseen categories. For genus transfer, unseen species of \textit{Quercus} achieve low test MAEs ($3.79$, $3.62$) by benefiting from related \textit{Quercus} training samples. In contrast, the taxonomically isolated \textit{Litchi chinensis} suffers a higher error ($15.61$) despite comparable density. Similarly, in cross-organ evaluation, \textit{Ricinus communis} (trained on leaves, tested on fruits) outperforms the isolated \textit{Momordica charantia} (MAE $4.58$ vs. $7.60$) under identical density.

While taxonomic priors offer useful guidance, visual cues remain critical. Evaluating the zero-shot approach GroundingREC~\cite{Dai_2024_CVPR} with taxonomic prompts yields a test MAE of $24.14$ and $R^2$ of $0.53$, falling behind visual-exemplar methods. Replacing LOCA's ResNet-$50$ backbone with BioCLIP$2$~\cite{gu2025bioclip} also yields inferior results (MAE $34.75$, $R^2$ $0.29$), likely because the low-resolution feature maps of ViT architectures without additional adapter designs are suboptimal for dense prediction. 
Since simple text encoding and off-the-shelf backbones underperform, moving beyond text to explicitly model inherent visual similarities among related species offers a promising path toward robust, fine-grained representations.

\section{Conclusion}
We present \datasetName{}, the first large-scale plant counting benchmark grounded in plant taxonomy. 
In contrast to prior counting datasets dominated by rigid or man-made objects, \datasetName{} captures the hierarchical structure, morphological diversity, and multi-scale complexity of real-world plants. 
To assess its utility, we benchmark several representative CAC models and show incorporating plant taxonomic information is crucial for robust and generalizable counting in natural scenes. We hope \datasetName{} will broaden the scope of visual counting research by challenging models to move beyond familiar regimes focused on rigid objects and by encouraging the development of representations to reason over fine grained structures and hierarchical organization.

For future work, we plan to extend \mbox{\datasetName{}} with additional plant species, richer temporal and environmental annotations, and more biological organizational levels such as the cellular level to support biologically grounded counting.

\vspace{20pt}
\noindent\textbf{Acknowledgement.} This work is supported in part by the National Natural Science Foundation of China under Grant No.~62576146 and in part by the HUST Undergraduate Natural Science Foundation under Grant No.~62500034.

{
    \small
    \bibliographystyle{ieeenat_fullname}
    \bibliography{main}
}

\clearpage
\setcounter{page}{1}
\maketitlesupplementary
\appendix

The supplementary material provides the algorithms for dataset partitioning, the method for mapping the hierarchical taxonomic structure, qualitative visualizations and some image examples. First, \textbf{Sec.~\ref{sec:milp}} details the Mixed-Integer Linear Programming (MILP) formulation for dataset partitioning. Second, \textbf{Sec.~\ref{sec:taxonomy}} explains how we construct the mapping from taxonomic names to hierarchical codes. Finally, \textbf{Sec.~\ref{sec:vis}} presents both qualitative comparisons with baselines and diverse visual examples.

\section{MILP-based Dataset Partitioning}
\label{sec:milp}

We propose a Mixed-Integer Linear Programming (MILP) approach to partition the dataset. This ensures taxonomic independence between subsets while maintaining statistical balance in instance density.

\subsection{Process Description}

The pipeline consists of three stages:

\vspace{4pt}
\noindent\textbf{Build Units.}
We group images into atomic units based on unique \textit{Species-Organization} pairs.
For example, all images of \textit{Zea mays-flower} form a single unit.
Each atomic unit corresponds to a unique visual phenotype of a single species
and is assigned exclusively to one subset.

\vspace{4pt}
\noindent\textbf{Optimize Assignments.}
The unit assignments are determined using the PULP\_CBC\_CMD solver.
The optimization objective combines two types of deviation penalties.
First, we minimize how far the image counts ($N_k$) deviate from the target
$7\!:\!1\!:\!2$ proportion, ensuring that the group sizes remain close to this
desired ratio.
Second, we minimize the imbalance in the average number of object instances
(points) ($P_k$) across groups.
Since matching the image ratio is more important, we assign it a much larger
weight ($\lambda_{img} = 100.0$) compared to the weight for balancing instance
counts ($\lambda_{pts} = 1.0$). This forces the solver to satisfy the image
ratio first, and only then adjust the instance distribution.

\vspace{4pt}
\noindent\textbf{Ensure Coverage.}
To ensure that the model is exposed to all levels of biological organization,
we enforce a hard constraint that the training split must include at least one
unit from every category. After the solver completes, we also verify that each
of the three subsets contains data from all observation scales (Microscopy,
Close-range, Remote Sensing). If any subset lacks a particular scale, the
partition is rejected and the optimization is restarted with a different random
seed for the solver’s internal heuristics, ensuring a new search trajectory and
a fresh candidate solution.

\subsection{Problem Formulation}

Let $\mathcal{D}$ be the full dataset containing images $I$, and let $\mathcal{U} = \{u_1, \dots, u_M\}$ denote the set of atomic units, where each unit corresponds to a unique \textit{Species-Organization} pair. For each unit $u$, we denote by $n_u$ the number of images and by $p_u$ the total number of object instances.

Let $\mathcal{O}$ be the set of biological organization categories, and let $\mathcal{U}_o \subseteq \mathcal{U}$ denote the units belonging to category $o$. Similarly, let $\mathcal{T}$ be the set of observation scales ($\text{Microscopy}$, $\text{Close-range}$, $\text{Remote Sensing}$), and let $\mathcal{U}_t \subseteq \mathcal{U}$ denote the units originating from scale $t$.

We partition $\mathcal{U}$ into three subsets $\mathcal{S} = \{\text{train}, \text{val}, \text{test}\}$ with target proportions $r_k = \{0.7, 0.1, 0.2\}$.

\paragraph{Decision Variables.}
For each unit $u$ and split $k$, we define a binary assignment variable:
\[
x_{u,k} =
\begin{cases}
1, & \text{if unit } u \text{ is assigned to split } k, \\
0, & \text{otherwise}.
\end{cases}
\]

\paragraph{Objective Function.}
We minimize the weighted deviation between the actual and target totals for image counts ($\hat{N}_k$) and point counts ($\hat{P}_k$). The objective function is formulated as:

\begin{equation}
\begin{split}
\min \sum_{k \in \mathcal{S}} \bigg( & \lambda_{\text{img}} \left| \sum_{u \in \mathcal{U}} n_u x_{u,k} - \hat{N}_k \right| \\
& + \lambda_{\text{pts}} \left| \sum_{u \in \mathcal{U}} p_u x_{u,k} - \hat{P}_k \right| \bigg)
\end{split}
\end{equation}
where $\hat{N}_k = r_k \sum_{u} n_u$ and $\hat{P}_k = r_k \sum_{u} p_u$. We use $\lambda_{\text{img}} = 100$ and $\lambda_{\text{pts}} = 1$ to strictly prioritize the adherence to the image-count ratio.

\paragraph{Constraints.}
The optimization is subject to the following constraints:

\begin{enumerate}[label=\roman*)]
    \item \textit{Unique assignment.} Each unit must be assigned to exactly one split:
    \begin{equation}
        \sum_{k \in \mathcal{S}} x_{u,k} = 1, \quad \forall u \in \mathcal{U}.
    \end{equation}

    \item \textit{Training set coverage.} To ensure robust feature learning, the training subset is explicitly constrained to include at least one unit from each organization category:
    \begin{equation}
        \sum_{u \in \mathcal{U}_o} x_{u,\text{train}} \ge 1, \quad \forall o \in \mathcal{O}.
    \end{equation}
\end{enumerate}

\paragraph{Verification.}
While the MILP guarantees organization coverage for the training set, we additionally require that all splits cover the full spectrum of observation scales. After optimization, we verify that:
\begin{equation}
    \sum_{u \in \mathcal{U}_t} x_{u,k} \ge 1, \quad \forall t \in \mathcal{T},\; \forall k \in \mathcal{S}.
\end{equation}
If this condition is not met for any split $k$, the resulting partition is discarded, and the optimization is re-initialized with a different random seed until a valid solution is obtained.

\section{Hierarchical Taxonomic Encoding}
\label{sec:taxonomy}



We follow the standard $7$-hierarchy biological taxonomy—\textit{Kingdom, Phylum, Class, Order, Family, Genus,} and \textit{Species}. To build the index mappings, we first sort all species names in the dataset and iterate through them in order. For each species, we retrieve its taxonomic labels across all remaining hierarchies. Whenever a hierarchy encounters a category name for the first time, we append it to that level’s dictionary and assign the next integer index starting from \(1\). Formally, for each hierarchy \(h\), this procedure defines a mapping
\[
    \phi_h : \mathcal{C}_h \rightarrow \{1, 2, \dots, |\mathcal{C}_h|\},
\]
where \(\mathcal{C}_h\) denotes the set of unique category names observed at hierarchy \(h\). This construction yields a complete mapping from taxonomic names to integer identifiers for every hierarchy.

Under this protocol, each species is encoded as a $7$-dimensional vector $v$:
$$ v = \left[ id_{\text{king}},\, id_{\text{phy}},\, id_{\text{cls}},\, id_{\text{ord}},\, id_{\text{fam}},\, id_{\text{gen}},\, id_{\text{spec}} \right] $$
where each component $id$ indicates the integer index of the taxon at that specific hierarchy.

\paragraph{Example.}
Consider the species \textit{Malus domestica} . Its taxonomic path maps to the following indices:

\noindent Kingdom: \textit{Plantae} $\to 1$; Phylum: \textit{Tracheophyta} $\to 1$; Class: \textit{Magnoliopsida} $\to 1$; Order: \textit{Rosales} $\to 14$; Family: \textit{Rosaceae} $\to 39$; Genus: \textit{Malus} $\to 113$; Species: \textit{Malus domestica} $\to 136$.

\noindent Consequently, the hierarchical encoding vector is:
$$ v_{\textit{Malus domestica}} = \left[ 1,\, 1,\, 1,\, 14,\, 39,\, 113,\, 136 \right] $$


The file \texttt{taxonomy\_ids.json} contains the complete index mappings for all categories across all $7$ taxonomic hierarchies. This file is included with the supplementary materials, and its structure is illustrated below:

\begin{verbatim}
{
  "Kingdom": {
    "Plantae": 1,
    "Fungi": 2
  },
  "Phylum": {
    "Tracheophyta": 1,
    "Basidiomycota": 2
  },
  ...
  "Family": {
    "Malvaceae": 1,
    "Fabaceae": 2,
    ...
    "Rhamnaceae": 83
  },
  ...
  "Species": {
    "Abelmoschus esculentus": 1,
    ...
    "Zea mays": 240,
    "Ziziphus mauritiana": 242
  }
}
\end{verbatim}

\section{Visualizations and Examples}
\label{sec:vis}

\begin{figure*}
    \centering
    \includegraphics[width=\linewidth]{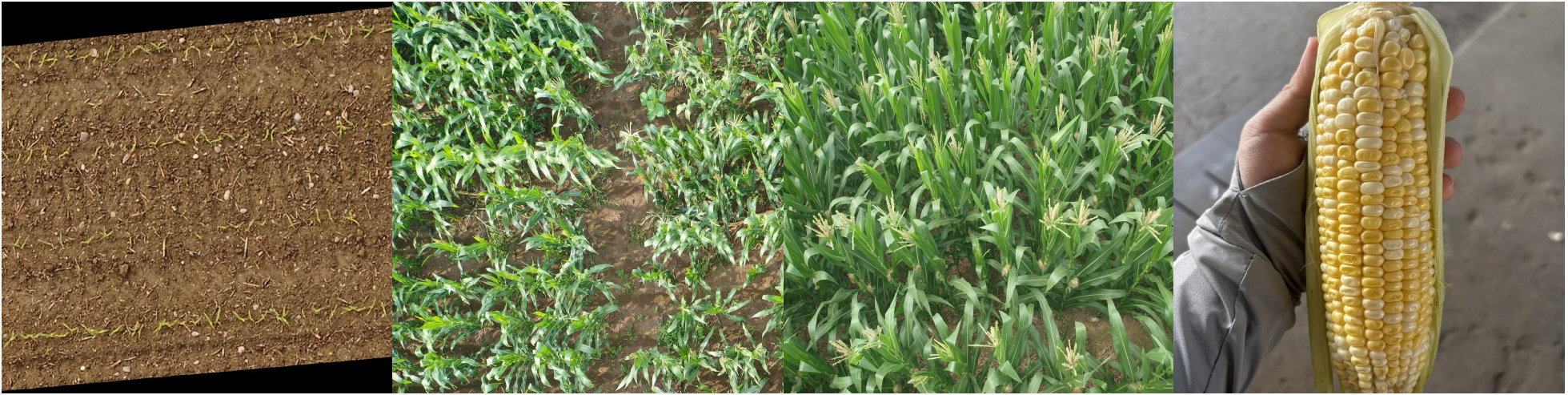}
    \caption{\textbf{Examples of the same species at different growth stages in our \datasetName{}.} The images span early seedling, vegetative development, tasseling, and final maturity.}
    \label{fig:stage}
\end{figure*}

Fig. \ref{fig:stage} shows images of a single species observed at different growth stages, where the changes in plant size, structure, and overall appearance become clearly visible across the timeline.

\begin{figure*}
    \centering
    \includegraphics[width=\linewidth]{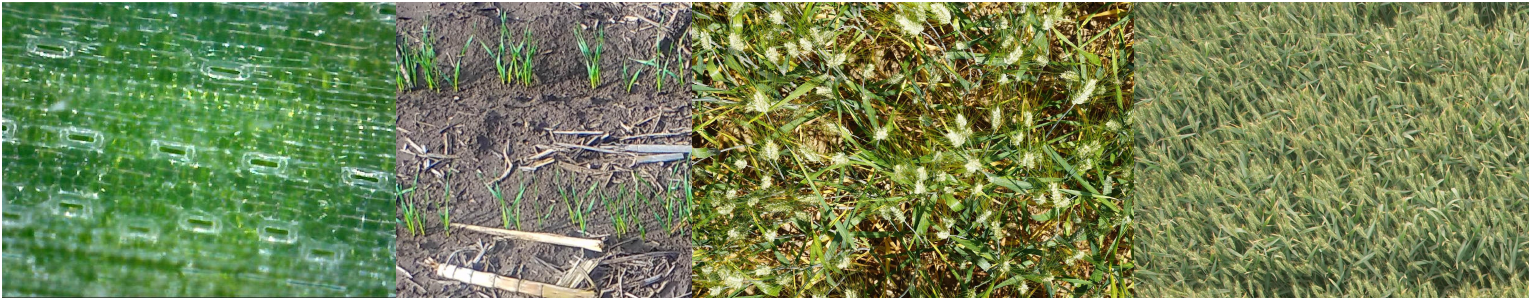}
    \caption{\textbf{Examples of different biological organizations from the same species in our \datasetName{}.} Tissue-level, organ-level, and whole-plant-level images show distinct texture patterns and structural characteristics.}
    \label{fig:org}
\end{figure*}

Fig. \ref{fig:org} presents examples from several biological organizations of the same species. The differences in texture, geometry, and visual scale across tissues, organs, and whole plants are evident from these samples.

\begin{figure*}
    \centering
    \includegraphics[width=\linewidth]{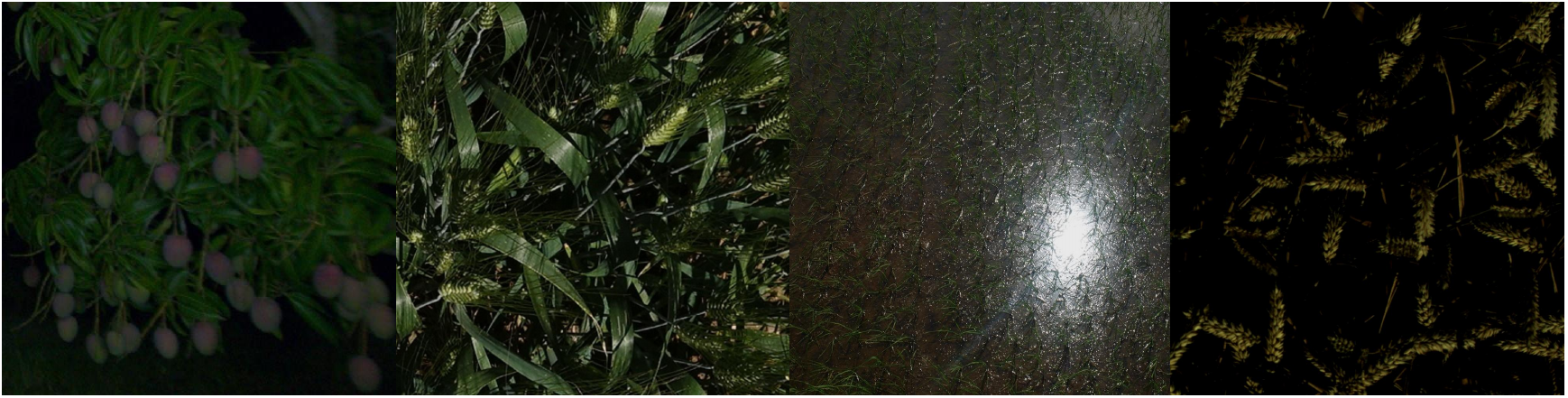}
    \caption{\textbf{Examples of images captured in dark or low-illumination environments from our \datasetName{}.} Reduced contrast, color distortion, and shadow-induced ambiguity complicate instance perception.}
    \label{fig:dark}
\end{figure*}

Fig. \ref{fig:dark} includes images collected under low-light conditions. Reduced visibility and unstable illumination make object boundaries harder to distinguish.

\begin{figure*}
    \centering
    \includegraphics[width=\linewidth]{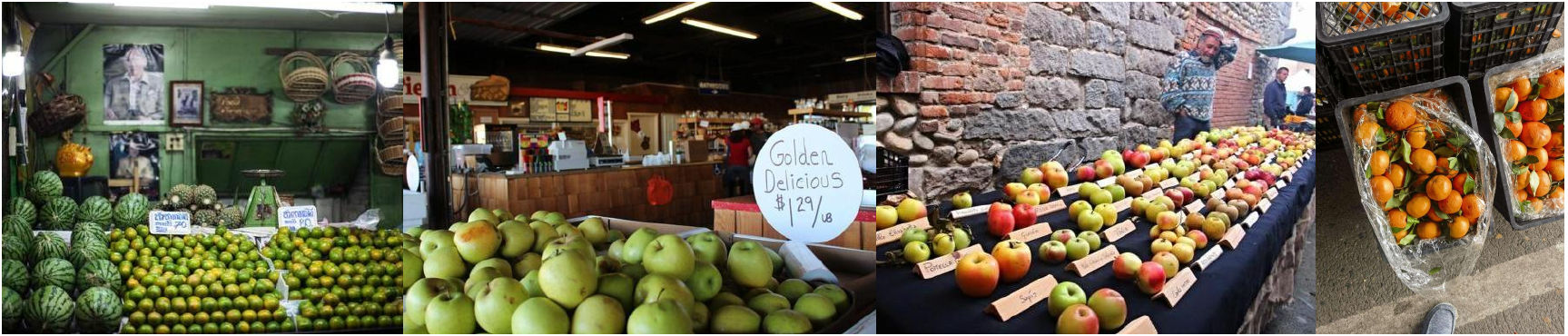}
    \caption{\textbf{Examples of real-world scenarios in our \datasetName{}.} Natural backgrounds, occluding structures, and cluttered surroundings reflect practical field conditions.}
    \label{fig:life}
\end{figure*}

Fig. \ref{fig:life} shows real-world scenes, with irregular backgrounds, occlusions, and surrounding environment adding substantial visual clutter.

\begin{figure*}
    \centering
    \includegraphics[width=\linewidth]{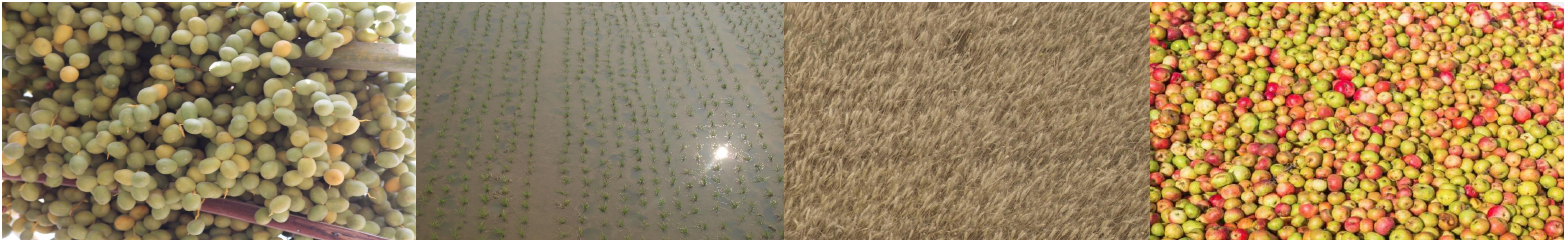}
    \caption{\textbf{Examples of high-density counting scenarios in our \datasetName{}.} These images include scenes containing hundreds or even thousands of instances, characterized by compact spatial arrangement, heavy overlap, and minimal inter-instance separation.}
    \label{fig:dense}
\end{figure*}

Fig. \ref{fig:dense} provides representative high-density cases containing large numbers of closely arranged instances. The crowded layouts and overlapping structures make accurate separation and counting particularly difficult.

\begin{figure*}[t]
    \centering
    \includegraphics[width=\linewidth]{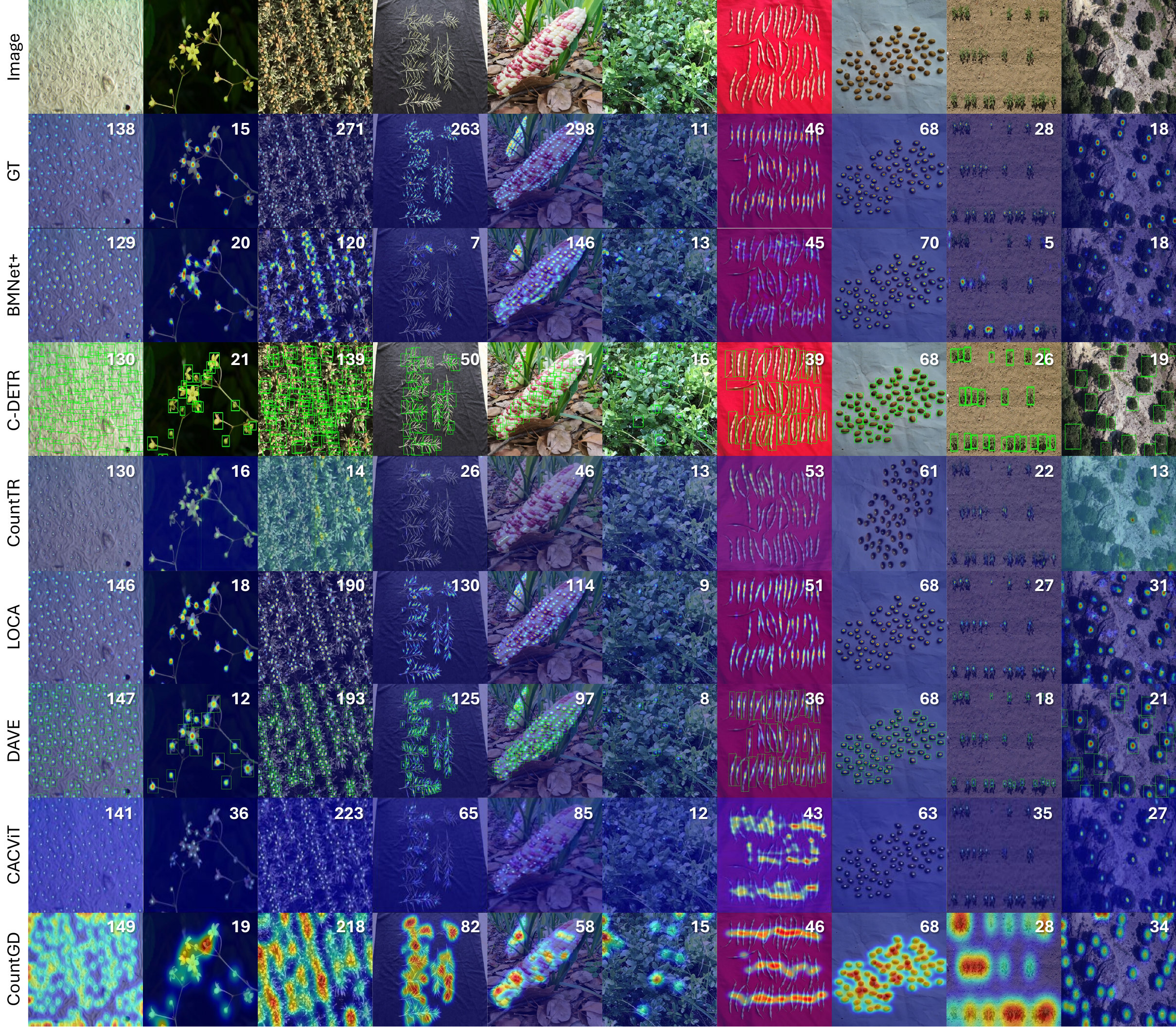}
    \caption{
    \textbf{Qualitative results of representative counting methods on our \datasetName{}.} The examples cover diverse plant forms, observation scales, and density levels.}
    \label{fig:experiment_visualization}
\end{figure*}

As shown in Fig. \ref{fig:experiment_visualization}, we include qualitative comparisons across multiple baseline methods to illustrate their performance under different plant-counting conditions.

\end{document}